%% file: main.tex
\documentclass{article}

\usepackage{svg}
\usepackage{enumitem}
\usepackage{xcolor}
\usepackage{tikz}
\usepackage{xspace}
\usetikzlibrary{patterns}

\usepackage[normalem]{ulem} %
\usepackage[textsize=footnotesize,textwidth=2.2cm,backgroundcolor=red!15,linecolor=red!60,bordercolor=red!60]{todonotes}

\PassOptionsToPackage{numbers}{natbib}

\usepackage[eandd, final]{neurips_2026}

\usepackage[utf8]{inputenc} %
\usepackage[T1]{fontenc}    %
\usepackage{hyperref}       %
\usepackage{url}            %
\usepackage{booktabs}       %
\usepackage{amsfonts}       %
\usepackage{nicefrac}       %
\usepackage{microtype}      %
\usepackage{xcolor}         %

\input{head/title}

\input{head/authors}

\input{flow/flow}

\begin{document}

\maketitle
\input{head/abstract}
\input{body/body}

\newpage
{\small
    \bibliographystyle{plainnat}
    \bibliography{references}
}

\newpage
\appendix
\section*{Appendix}
\input{body/appendix/appendix}

\newpage
\input{checklist.tex}

\end{document}

%% file: head/title.tex
\title{Redefining Instance Matching:
\\A Unified Framework for Part-Aware Matching in Panoptic Segmentation Evaluation}%

%% file: head/authors.tex
\author{%
\textbf{Erik Großkopf}$^{1,*}$, \textbf{Soumya Snigdha Kundu}$^{2,*}$, \textbf{Hendrik Möller}$^{4,5}$, \textbf{Nicolas Münster}$^{1}$,\\
\textbf{Mehdi Astaraki}$^{8}$, \textbf{Paula Tamara Buzduga}$^{1}$, \textbf{Kerstin Ritter}$^{1}$, \textbf{Benedikt Wiestler}$^{6}$,\\
\textbf{Jan Kirschke}$^{6}$, \textbf{Jonathan Shapey}$^{2,3}$, \textbf{Tom Vercauteren}$^{2}$, \textbf{Florian Kofler}$^{1,7,9,10}$\\[1ex]
$^{1}$Hertie Institute for AI in Brain Health, University of T\"ubingen, T\"ubingen, Germany\\
$^{2}$King's College London, London, UK\\
$^{3}$King's College Hospital, London, UK\\
$^{4}$Department of Neuroradiology, TUM University Hospital, Munich, Germany\\
$^{5}$Chair for AI in Healthcare and Medicine,\\
Technical University of Munich (TUM), Munich, Germany\\
$^{6}$Technical University of Munich, Munich, Germany\\
$^{7}$AI for Image-Guided Diagnosis and Therapy, TUM School of Medicine and Health,\\
Technical University of Munich, Munich, Germany\\
$^{8}$Stockholm University, Stockholm, Sweden\\
$^{9}$Department of Quantitative Biomedicine, University of Zurich, Zurich, Switzerland\\
$^{10}$Helmholtz AI, Helmholtz Munich, Neuherberg, Germany\\[1ex]
$^{*}$Equal contribution\\
\texttt{erik.grosskopf@uni-tuebingen.de}
}

%% file: flow/flow.tex
\usepackage{float}

\usepackage{graphicx}

\usepackage{amsmath}
\usepackage{amsthm}

\input{flow/acronyms}

\input{flow/commands}
\usepackage{hyperref}
\hypersetup{
  colorlinks=true,
  urlcolor=blue,
  linkcolor=black,
  citecolor=black,
}

\usepackage{varioref}

\usepackage{cleveref}

\usepackage{tabularx}

\usepackage{booktabs}

\usepackage{csquotes}

%% file: flow/acronyms.tex
\usepackage[nolist]{acronym}

\newacro{TP}{true positive}
\newacro{FP}{false positive}
\newacro{FN}{false negative}
\newacro{OS}{oversegmentation}
\newacro{US}{undersegmentation}
\newacro{IOU}[IoU]{intersection over union}
\newacro{PS}{panoptic segmentation}

\newacro{PQ}{Panoptic Quality}
\newacro{SQ}{Segmentation Quality}
\newacro{RQ}{Recognition Quality}
\newacro{NPQ}{Normalized Panoptic Quality}
\newacro{AUTC}{Area under Threshold Curve}

\newacroindefinite{FP}{an}{a}
\newacroindefinite{FN}{an}{a}
\newacroindefinite{OS}{an}{an}
\newacroindefinite{US}{a}{an}
\newacroindefinite{IOU}{an}{an}
\newacroindefinite{SQ}{an}{a}
\newacroindefinite{RQ}{an}{a}
\newacroindefinite{NPQ}{an}{a}
\newacroindefinite{AUTC}{an}{an}

%% file: flow/commands.tex
\newcommand{\TP}{\mathit{TP}}
\newcommand{\FN}{\mathit{FN}}
\newcommand{\FP}{\mathit{FP}}
\newcommand{\IoU}{\text{IoU}}
\newcommand{\BN}{\mathbb N}
\newcommand{\SL}{{\cal L}}
\newcommand{\tss}[1]{\textsuperscript{#1}}
\newcommand{\things}{\tss{Th}\xspace}
\newcommand{\stuff}{\tss{St}\xspace}

\DeclareMathOperator*{\argmax}{arg\,max}

%% file: head/abstract.tex
\begin{abstract}
The Panoptic Quality metric is the standard for jointly evaluating instance and semantic segmentation.
Its original definition matches predicted and ground truth segments only if their intersection over union exceeds 0.5, a threshold that guarantees a One-to-One matching that is both optimal and unique.
Below this threshold, the uniqueness guarantee is lost and multiple matching strategies emerge in a poorly explored problem space.
We systematically elucidate this space by recasting segment matching as a constrained bipartite assignment problem.
Independently bounding the prediction- and ground-truth-side degrees yields four matching strategies: One-to-One, Many-to-One, One-to-Many, and Many-to-Many.
We show that the first three yield interpretable instance-level assignments, whereas Many-to-Many entangles over- and under-segmentation and admits no consistent attribution of errors.
These strategies become relevant when instances are fragmented, adjacent objects are difficult to delineate, or annotations are noisy.
Central to our framework is a \emph{vertex-based} accounting of TP, FP, and FN, anchored to ground truth and predicted segments rather than to matching edges.
We further show that the framework extends naturally to part-aware panoptic segmentation and illustrate this extension conceptually on biomedical examples.
Across configurable case studies we report how different combinations of thresholds and matching strategies behave in practice.
\end{abstract}

%% file: body/body.tex
\input{body/introduction/intro}
\input{body/preliminaries/preliminaries}
\input{body/related_work/related_work}

\input{body/panoptica/panoptica}
\input{body/experiments/experiments}
\input{body/discussion/discussion}
\input{body/acknowledgements/acknowledgements}

%% file: body/introduction/intro.tex
\Ac{PS}~\citep{kirillov2019panoptic} unifies semantic and instance segmentation under a single metric, \ac{PQ}.
\ac{PQ} counts a predicted and a ground truth segment as a match only if their \ac{IOU} \textit{strictly exceeds} a fixed 0.5 threshold.
Because segments within a panoptic output do not overlap, at most one prediction can exceed this threshold for a given ground truth segment, and vice versa, so the matching is One-to-One by construction and reduces to a lookup.
Below 0.5, several candidates can exceed the threshold, and the original definition gives no rule for choosing among them.
Two limitations follow.
First, the One-to-One rule is overly restrictive whenever predictions and references legitimately split or merge.
An instance correctly recovered as several adjacent fragments yields at most one \ac{TP}, with every remaining fragment counted as \iac{FP}, even if their union matches the reference well.
Several recent metrics tackle parts of this problem.
Examples include proximity-based per-component evaluation~\citep{jaus2025every}, topology-aware connected-component Dice~\citep{rouge2025ccdice}, and many-to-many Cluster Dice~\citep{kundu2025cluster}.
Each one targets a single failure mode.
They do not share a common framework, and none extends to hierarchical evaluation.
Second, the 0.5 threshold is an opinionated default.
A prediction with $\mathrm{IoU} = 0.5$ is treated identically to one with zero overlap, and the choice of threshold heavily shapes the resulting score.
These issues particularly affect scenarios where instances are small, dense, or hierarchically organized: tiny-instance benchmarks, where a one-pixel shift dominates \ac{IOU}~\citep{foucart2023panoptic}, scene parsing with internal part annotations~\citep{degeus2021partaware}, and biomedical data, where lesions decompose into sub-regions and where expert disagreement induces ambiguous instance boundaries.
We address both limitations by recasting segment matching as a weighted bipartite assignment problem~\citep{chazalon2021revisiting}.
Edges are formed between predicted and ground truth segments with \ac{IOU} above a configurable threshold $\tau \in [0, 1)$, and \acp{TP}, \acp{FP}, and \acp{FN} are anchored to the vertices of the resulting graph rather than to matched edges.
This vertex-centric formulation preserves the invariant $|\TP| + |\FN| = |G|$ across all matching strategies.
It admits four strategies: One-to-One, Many-to-One, One-to-Many, and Many-to-Many.
The first three are well-defined within the \ac{PQ} framework, and each one either tolerates or penalizes a specific segmentation failure mode.
Many-to-Many loses this interpretability and falls outside the framework.
At $\tau = 0.5$, the One-to-One case recovers the original \ac{PQ} exactly.
The same formalism defines the \ac{AUTC}, the threshold-agnostic summary $\int_0^1 \mathrm{PQ}(\tau)\,d\tau$, which inherits the same configurability across strategies and parts.
\ac{AUTC} decomposes into segmentation- and recognition-specific terms that isolate the source of a method's threshold-averaged performance.
For settings with dense, touching instances, we additionally describe a Voronoi-based region-wise variant.
It complements One-to-Many matching by partitioning the image into per-instance cells, in which the same bipartite construction is applied locally.
\\
Part-aware Panoptic Segmentation~\citep{degeus2021partaware} extended \ac{PQ} to hierarchical targets, where a scene-level instance carries a meaningful part decomposition, but inherited the matching rule unchanged.
As an add-on for hierarchical targets, we extend PartPQ to this framework and decouple it from the original One-to-One rule.
The part-aware quality term then composes with any of our matching strategies.
Finally, we assess the practical consequences of these choices on COCO and in controlled perturbation studies, and illustrate the part-aware extension on biomedical examples in accompanying notebooks.
We implement the One-to-One, Many-to-One and One-to-Many strategies, Voronoi-based region-wise analysis, part-aware evaluation, and Area Under Threshold Curve computation as configurable extensions to the open-source \href{https://github.com/BrainLesion/panoptica}{Panoptica} package.
The package also integrates the prior approaches discussed above as configurable options, so practitioners can select the matcher and quality term that fits their task and requirements rather than adopting a tool's default.

%% file: body/preliminaries/preliminaries.tex
\section{Preliminaries}
\label{sec:preliminaries}

Semantic segmentation assigns every pixel/voxel $i$ a single class label $l_i \in \SL$~\citep{long2015fully}.
Pixels/voxels are grouped only by class, so distinct instances of the same class merge into one region.
Instance segmentation requires a method to delineate every object instance in an image with a binary mask, typically accompanied by a class label and a confidence score~\citep{he2017mask,hariharan2014simultaneous}.
Unlike \ac{PS}, it considers only instance labels and permits overlapping masks, since each instance is predicted independently rather than as a partition of the image.
\Ac{PS}, introduced by \citet{kirillov2019panoptic}, unifies semantic and instance segmentation into a single coherent task.
Given a set of $L$ semantic classes $\SL = \SL\stuff \cup \SL\things$, \iac{PS} algorithm assigns every pixel/voxel $i$ a pair $(l_i, z_i) \in \SL \times \BN$, where $l_i$ denotes the semantic class and $z_i$ the instance id.
\emph{Stuff} classes (e.g.\ \emph{grass}, \emph{sky}) are amorphous regions for which the instance id is irrelevant, while \emph{thing} classes (e.g.\ \emph{sheep}, \emph{car}) are countable objects whose pixels/voxels must be partitioned into distinct instances.
Unlike instance segmentation, \ac{PS} forbids overlapping segments by construction, so each pixel/voxel receives exactly one label pair (see visual comparison of segmentation tasks in \cref{sec:appendix_figures}).

\paragraph{Panoptic Quality.}
To evaluate \ac{PS} on a single scale, \citet{kirillov2019panoptic} introduced the \ac{PQ} metric.
\ac{PQ} proceeds in two steps: (i) matching predicted to ground truth segments, and (ii) aggregating the matches into a single score per class, which is then averaged over classes.
A predicted segment $p$ matches a ground truth segment $g$ when their intersection-over-union $\IoU(p,g) = |p \cap g| / |p \cup g|$ satisfies $\IoU(p, g) > 0.5$.
Together with the non-overlap property of \ac{PS}, this threshold guarantees a \emph{unique} matching: at most one predicted segment can match each ground truth segment, and vice versa, so no assignment problem needs to be solved~\citep[Theorem~1]{kirillov2019panoptic}.
The matched pairs form the \acp{TP}, while unmatched predictions and ground truth segments form the \acp{FP} and \acp{FN}, respectively. 
\ac{PQ} is then defined as
\begin{equation}\label{eq:pq}
\text{PQ}
= \frac{\sum_{(p,g) \in \TP} \IoU(p,g)}{|\TP| + \tfrac{1}{2}|\FP| + \tfrac{1}{2}|\FN|}
= \underbrace{\frac{\sum_{(p,g) \in \TP} \IoU(p,g)}{|\TP|}}_{\text{SQ}}
\times \underbrace{\frac{|\TP|}{|\TP| + \tfrac{1}{2}|\FP| + \tfrac{1}{2}|\FN|}}_{\text{RQ}},
\end{equation}

which decomposes into \iac{SQ} term, the average \ac{IOU} of matched segments, and \iac{RQ} term, equivalent to the $F_1$ score~\citep{vanrijsbergen1979information}. All segments contribute equally regardless of area.

\paragraph{Part-aware Panoptic Quality.}
\citet{degeus2021partaware} extend \ac{PS} with a third level of granularity: each thing instance may carry an internal decomposition into a fixed set of \emph{parts}, e.g.\ a \emph{person} into head, torso, arms, and legs.
The corresponding metric, Part-aware Panoptic Quality (\textit{PartPQ}), inherits the matching machinery of \ac{PQ} and reweights only the per-pair quality so that parts contribute to the score only through a successfully matched parent instance.
To our knowledge, PartPQ is the only existing panoptic-style metric that jointly evaluates instances and their internal parts, and it has not previously been applied to biomedical data.
We revisit it formally in \cref{sec:part-aware} and showcase its adoption within the medical image analysis framework.

\paragraph{The 0.5 threshold.}
\citet{kirillov2019panoptic} argue that lower thresholds are unnecessary because matches with $\IoU \le 0.5$ are rare in practice.
This assumption breaks down in domains where boundary localization is inherently ambiguous or where small, densely packed instances dominate the data.
Biomedical imaging offers a plethora of such regimes, including multiple sclerosis lesion segmentation~\citep{kofler2023blob} and brain-metastasis segmentation in MRI~\citep{maleki2025analysis}, liver and abdominal metastasis in CT~\citep{bilic2023liver,heller2021state,bassi2025radgpt}, encapsulin segmentation in electron microscopy~\citep{sigmund2023genetically}, and neuron segmentation in light-sheet microscopy~\citep{kaltenecker2024virtual}.
In clinical scenarios such as liver metastasis or head and neck cancer, multiple independent lesions appear in close proximity within a restricted anatomical volume, yet must be identified as separate entities to monitor disease progression and treatment response.
Under the One-to-One rule, otherwise correctly predicted segments in such regimes are systematically penalized as simultaneous \acp{FP} and \acp{FN}, obscuring real model improvements. 

%% file: body/related_work/related_work.tex
\section{Related Work} \label{sec:related_work}
The Metrics Reloaded framework recommends \ac{PQ} for the task of instance segmentation as it fuses both detection performance and segmentation quality of successfully matched segments into a single score \citep{maier-hein2024metrics}. 
While the authors discuss common pitfalls, such as class imbalance, they do not offer any means to address these difficulties for instance segmentation evaluation.
\\
Defining a fixed matching threshold to classify predictions as either \acp{TP} or \acp{FP} results in a high sensitivity to the choice of the threshold. 
Predictions overlapping reference instances just slightly below or equal to the threshold are disregarded completely and penalized harshly.
\citet{chen2023sortedap} introduced sortedAP in order to completely omit a fixed threshold by instead integrating over all existing \ac{IOU} thresholds.
To overcome the limitation of a threshold strictly greater than 0.5, the authors propose using the Hungarian algorithm for matching, which preserves a One-to-One relationship between segments and maximizes the accumulated \acp{IOU} of matches.
\\
Motivated by a bias towards larger instances in semantic segmentation evaluation of metastases, \citet{jaus2025every} proposed Connected-Component Metrics incorporating per-component evaluation and a proximity-based matching criterion, giving each tumor the same weight irrespective of its size.
\\
\citet*{chazalon2021revisiting} reformulate the matching of predicted and ground truth segments as an assignment problem on a bipartite graph in the context of historical map evaluation. 
They define a matching as a One-to-One relation between the disjoint vertex sets of predictions and ground truth segments, with unmatched vertices allowed on either side. 
The same work introduces \ac{NPQ}, which integrates the F-score over matching thresholds and is argued to be an equivalent rewriting of the standard \ac{PQ} restricted to $\tau = 0.5$. 
We adopt the bipartite formulation and extend it to sub-$0.5$ thresholds and different matching strategies (\cref{sec:methods}). 
The relation between \ac{NPQ}, sortedAP, and \ac{AUTC} is discussed in \cref{sec:autc}.

%% file: body/panoptica/panoptica.tex
\input{body/panoptica/fig_matching}

\section{Rethinking Segment Matching for Panoptic Quality} \label{sec:methods}
We recast the matching of predicted to ground truth segments as a constrained bipartite assignment problem~\citep*{chazalon2021revisiting} to admit well-defined extensions for $\tau < 0.5$.
Let $G = \{g_1, \ldots, g_m\}$ and $P = \{p_1, \ldots, p_n\}$ denote the ground truth and predicted segments of a given class, with $G \cup P \neq \emptyset$, and let $\tau \in [0, 1)$ be a matching threshold.
We construct a weighted bipartite graph $\mathcal{B}_\tau = (G \cup P,\, E_\tau,\, w)$ with edges corresponding to overlapping pairs of prediction and ground truth segments whose \ac{IOU} is greater than the threshold $\tau$
\begin{equation}\label{eq:edges}
    E_\tau = \{(g_i, p_j) \in G \times P : \IoU(g_i, p_j) > \tau\}
\end{equation}

and edge weights $w(g_i, p_j) = \IoU(g_i, p_j)$.
A \emph{matching} $M \subseteq E_\tau$ is a subset of these edges. 
Let $\deg_M(v)$ denote the number of edges in $M$ incident to vertex $v$. 
A matching strategy is defined by independently constraining $\deg_M$ on each side of $\mathcal{B}_\tau$ to at most one or leaving it unconstrained.
The first word of a strategy's name refers to the prediction and the second to the ground truth side: \textit{Many-to-One} therefore means $\deg_M(p) \leq 1$ while $\deg_M(g)$ is unconstrained, allowing multiple predicted segments to match a single ground truth segment (\cref{fig:matching-comparison} (b)).
The four resulting strategies are treated in detail in \cref{sec:one-to-one,sec:many-to-one,sec:one-to-many,sec:many-to-many} and summarized in \cref{tab:matching-strategies}.
The discussion of the optimization of $M$ under each strategy is consolidated in \cref{sec:optimal-matching}.

\paragraph{Vertex-based TP, FP, and FN.}
In the original \ac{PQ} formulation, the One-to-One construction makes edges and matched vertices interchangeable: each edge in $M$ corresponds to exactly one matched ground truth and one matched predicted segment, and \acp{TP} can be identified with $M$ itself or with the matched ground truth set.
Once we allow thresholds smaller than 0.5 and admit Many-to-One or One-to-Many matchings, the equivalence breaks: a single ground truth or predicted segment may be incident to multiple matched edges, and $|M|$ can exceed $|G|$ or $|P|$. We must therefore decide whether to count edges or vertices.
We anchor \ac{TP} and \ac{FN} to ground truth vertices and \ac{FP} to predicted vertices:
\begin{equation}\label{eq:tpfnfp}
    \TP = \{g \in G \!:\! \deg_M(g) \geq 1\}, \;
    \FN = \{g \in G \!:\! \deg_M(g) = 0\}, \;
    \FP = \{p \in P \!:\! \deg_M(p) = 0\}.
\end{equation}
\\
This vertex-centric formulation guarantees the invariant $|\TP|+|\FN|=|G|$ across all matching strategies, so \ac{TP} and \ac{FN} partition the ground truth set: every ground truth segment is either recognized or missed, never counted multiple times.
\ac{RQ} therefore retains its interpretation as the recognition rate of ground truth segments rather than over-matching edges: a ground truth segment jointly described by several predicted segments in a Many-to-One matching contributes one \ac{TP}, not several. 
\ac{SQ} remains an average \ac{IOU} per recognized ground truth segment, with each instance weighted equally regardless of how many predicted segments are matched to it.
The edge-based alternative $|\TP|=|M|$ would violate both properties under asymmetric matchings, where $|M|$ can exceed $|G|$ or $|P|$.

\paragraph{Segmentation quality under multi-edge matchings.}
For each matched ground truth segment $g \in \TP$, let $P_M(g) = \bigcup \{\, p \in P : (g,p) \in M \,\}$ denote the union of predicted segments matched to $g$. 
We then define
\begin{equation}\label{eq:sq}
    \text{SQ} = \frac{\sum_{g \in \TP} \IoU(P_M(g), g)}{|\TP|},
\end{equation}
which preserves the $[0,1]$ bound on \ac{SQ} and weights each recognized ground truth segment equally.
In One-to-One and One-to-Many matchings, $P_M(g)$ is always a single predicted segment and \cref{eq:sq} reduces to the standard per-edge form. 
The construction is only nontrivial in the Many-to-One case (\cref{sec:many-to-one}).

\begin{table}[tbp]
\centering
\caption{Matching strategies defined by bipartite degree constraints. The original \ac{PQ} One-to-One matching constrains both predictions and ground truth segments to be part of at most one match. Many-to-One and One-to-Many each relax one constraint. }
\label{tab:matching-strategies}
\begin{tabular}{@{}lccl@{}}
\toprule
Strategy & $\deg_M(p)$ & $\deg_M(g)$ & Interpretation \\
\midrule
One-to-One    & $\leq 1$ & $\leq 1$ & Each $p$ matches at most one $g$ and vice versa. \\
Many-to-One   & $\leq 1$ & $ $      & Multiple $p$ may jointly describe one $g$. \\
One-to-Many   & $ $      & $\leq 1$ & A single $p$ may cover multiple $g$. \\
Many-to-Many  & $ $      & $ $      & No partition structure; outside the \ac{PQ} framework. \\
\bottomrule
\end{tabular}
\end{table}

\subsection{One-to-One Matching} \label{sec:one-to-one}
In One-to-One matching, every predicted segment matches at most one ground truth segment and vice versa.
Formally, $M \subseteq E_\tau$ is a One-to-One matching if
\begin{equation}\label{eq:1to1}
    \deg_M(v) \leq 1 \quad \text{for all } v \in G \cup P.
\end{equation}
One-to-One matching credits exactly one predicted segment per ground truth segment, as depicted in \cref{fig:matching-comparison} (a).
If two predicted segments both overlap the same object, only the higher-\ac{IOU} one contributes to \iac{TP} while the other is counted as \iac{FP} (\cref{fig:matching-comparison} (b), left).
When $\tau = 0.5$, One-to-One corresponds to the original \ac{PQ} definition of \citet{kirillov2019panoptic}.
Like standard \ac{PQ}, this strategy penalizes \ac{OS} and \ac{US} symmetrically, and is the only strategy that does so.
If both failure modes carry equal cost in the downstream task, only One-to-One reflects that. 
This approach is well-suited for scenarios in which an exact assignment of predicted to ground truth segments is non-negotiable and the ground truth annotations exhibit low ambiguity and low noise.

\subsection{Many-to-One Matching} \label{sec:many-to-one}

In Many-to-One matching, multiple predicted segments may jointly describe a single ground truth segment, while each predicted segment is still matched to at most one ground truth segment.
Formally,
\begin{equation}\label{eq:mto1}
    \deg_M(p) \leq 1 \quad \text{for all } p \in P.
\end{equation}

Multiple predicted segments matched to a single ground truth segment are counted as a single \ac{TP} and reduce the total \ac{FP} count by the number of additional predicted segments absorbed into that match (\cref{fig:matching-comparison} (b)).
This setting is appropriate when \ac{OS} reflects a legitimate decomposition of a ground truth segment rather than a model failure, for example, when a long, thin instance is correctly recovered as several adjacent fragments.

\subsection{One-to-Many Matching} \label{sec:one-to-many}
In One-to-Many matching, a single predicted segment may describe multiple ground truth segments, while each ground truth segment is still matched by at most one predicted segment.
Formally,
\begin{equation}\label{eq:1tom}
    \deg_M(g) \leq 1 \quad \text{for all } g \in G.
\end{equation}

This setting is appropriate when the delineation of adjacent ground truth segments is ambiguous or a single predicted segment describing multiple references should still be credited rather than penalized as both \ac{FP} and \ac{FN}.
For example, individual cells in a dense cluster may not be separated in the predicted segment, yet still convey the correct semantic and approximate spatial extent.

\subsection{Many-to-Many Matching} \label{sec:many-to-many}
A Many-to-Many strategy drops the degree constraints on both sides of $\mathcal{B}_\tau$, so any $M \subseteq E_\tau$ is admissible.
The accounting of \cref{eq:tpfnfp,eq:sq} remains well-defined and the invariant $|\TP| + |\FN| = |G|$ still holds.
The difficulty lies in what the additional edges express.
Under each of the three preceding strategies, every connected component of $M$ is a star whose center lies either in $G$ or $P$, making it interpretable.
In Many-to-One, the center is a ground truth segment described by several predicted fragments.
In One-to-Many, it is a predicted segment that merges several ground truth segments and in One-to-One, every component is a single edge.
Without constraints, $M$ may also contain components that relate at least two predicted segments to at least two ground truth segments.
Every such component exhibits both \ac{OS} and \ac{US}, so no direction of attribution, from predictions to ground truth or vice versa, is consistent with it.
We therefore do not consider Many-to-Many matchings where predictions and references are related through such set-to-set structures.
The evaluation is better framed as a comparison of partitions than as an instance assignment.

\subsection{Area under Threshold Curve (AUTC)} \label{sec:autc}
\input{body/panoptica/fig_autc}

Choosing a fixed threshold strongly influences the evaluation by making an opinionated assumption about the minimum \ac{IOU} required for \iac{TP}.
And predicted segments which overlap a ground truth segment just slightly below the threshold are harshly penalized in the final \ac{PQ} computation.
We bypass this choice by treating $\tau$ as a free parameter and integrating \ac{PQ} over all thresholds, yielding a threshold-agnostic summary.
Let $\mathrm{PQ}(\tau)$ denote the \ac{PQ} computed using the optimal matching $M^\star_\tau$ derived from the thresholded bipartite graph $\mathcal{B}_\tau$. 
We define the \ac{AUTC} as
\begin{equation}
    \mathrm{AUTC} = \int_{0}^{1} \mathrm{PQ}(\tau) \, d\tau.
\end{equation}
Because \ac{PQ} itself takes values in $[0, 1]$ and we integrate it over the threshold parameter of unit length, \ac{AUTC} also lies in $[0, 1]$. Although defined as a continuous integral, \ac{AUTC} reduces to an exact finite sum. 
Let $E_0 = \{(g_i, p_j) \in G \times P : \IoU(g_i, p_j) > 0\}$ denote the edges with strictly positive overlap, and let $u_1 < u_2 < \dots < u_k$ be the unique edge weights $\{w(e) : e \in E_0\}$ in increasing order. 
The edge set $E_\tau$ changes only when $\tau$ crosses an element of this set, so $\mathrm{PQ}(\tau)$ is a right-continuous step function on $[0, 1)$.
Setting $u_0 = 0$, the integral becomes
\begin{equation}\label{eq:autc-discrete}
    \mathrm{AUTC} = \sum_{i=0}^{k-1} \mathrm{PQ}(u_i) \cdot (u_{i+1} - u_i),
\end{equation}
where the contribution from $\tau \geq u_k$ vanishes: no edge satisfies $\IoU > u_k$, so every $g \in G$ is \iac{FN}, every $p \in P$ is \iac{FP}, and $\mathrm{PQ}(\tau) = 0$ on $[u_k, 1)$.

\paragraph{Decomposing AUTC.}
Both factors of \ac{PQ} are themselves step functions in $\tau$, so the same construction yields recognition- and segmentation-specific summaries:
\begin{equation}
    \mathrm{AUTC}_\mathrm{SQ} = \int_{0}^{1} \mathrm{SQ}(\tau) \, d\tau,
    \qquad
    \mathrm{AUTC}_\mathrm{RQ} = \int_{0}^{1} \mathrm{RQ}(\tau) \, d\tau,
\end{equation}
both bounded in $[0, 1]$ by the same argument and computable via the discrete form of \cref{eq:autc-discrete}.
Since \ac{SQ} averages over $\TP$, it is undefined on $[u_k, 1)$, where $\TP = \emptyset$.
We set $\mathrm{SQ}(\tau) := 0$ there, which preserves $\mathrm{PQ}(\tau) = \mathrm{SQ}(\tau) \cdot \mathrm{RQ}(\tau)$.
As a result, $\mathrm{AUTC}_\mathrm{SQ}$ also reflects the threshold range over which any segment is recognized, not only the quality of recognized segments.
Reporting them alongside \ac{AUTC} isolates whether a method's threshold-averaged advantage stems from better segmentation quality ($\mathrm{AUTC}_\mathrm{SQ}$) or from a higher recognition rate across thresholds ($\mathrm{AUTC}_\mathrm{RQ}$).
We emphasize, however, that the multiplicative structure of \ac{PQ} does \emph{not} carry over to the integrals: in general
\begin{equation}
    \mathrm{AUTC} \;\neq\; \mathrm{AUTC}_\mathrm{SQ} \cdot \mathrm{AUTC}_\mathrm{RQ},
\end{equation}
because the integral of a product is not the product of integrals. 

\paragraph{Relation to existing threshold-agnostic metrics.}
Two prior metrics also integrate a panoptic-style score over matching thresholds.
\citet*{chazalon2021revisiting} introduce \ac{NPQ}, the area under the F-score curve over $\tau \in [0, 1)$ under One-to-One matching, and argue that the standard \ac{PQ} is an equivalent rewriting of \ac{NPQ} at $\tau \geq 0.5$.
\citet{chen2023sortedap} propose sortedAP, which integrates the average precision score over the full threshold range using Hungarian One-to-One matching. 
Both metrics are restricted to One-to-One matching.
In contrast, \ac{AUTC} additionally admits Many-to-One and One-to-Many matching strategies and decomposes into $\mathrm{AUTC}_{\mathrm{SQ}}$ and $\mathrm{AUTC}_{\mathrm{RQ}}$.

\begin{figure}
    \centering
    \begin{minipage}[b]{0.22\linewidth}
        \centering
        \includegraphics[width=\linewidth, height=3.2cm, keepaspectratio]{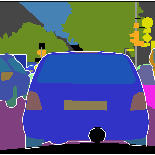}\\
        {\footnotesize (a) Cityscapes}
    \end{minipage}\hspace{0.02\linewidth}%
    \begin{minipage}[b]{0.22\linewidth}
        \centering
        \includegraphics[width=\linewidth, height=3.2cm, keepaspectratio]{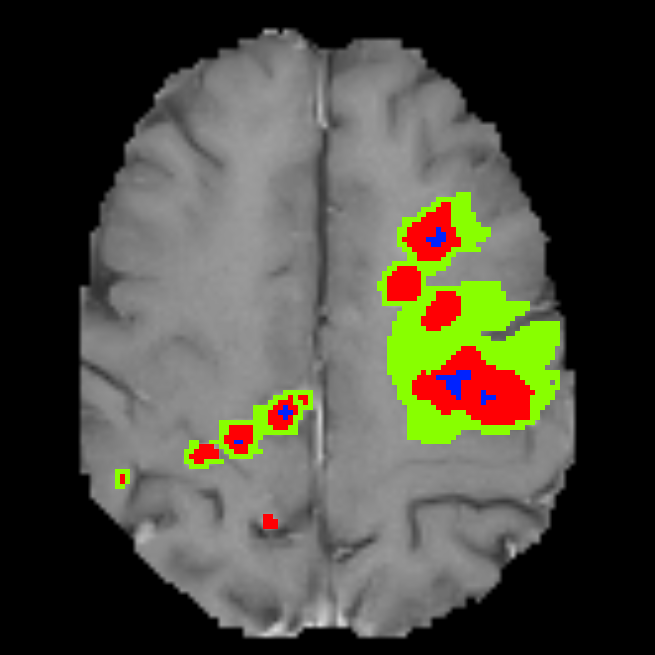}\\
        {\footnotesize (b) BraTS-METS2025}
    \end{minipage}\hspace{0.02\linewidth}%
    \begin{minipage}[b]{0.22\linewidth}
        \centering
        \includegraphics[width=\linewidth, height=3.2cm, keepaspectratio]{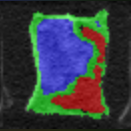}\\
        {\footnotesize (c) Vertebrae}
    \end{minipage}\hspace{0.02\linewidth}%
    \begin{minipage}[b]{0.22\linewidth}
        \centering
        \includegraphics[width=\linewidth, height=3.2cm, keepaspectratio]{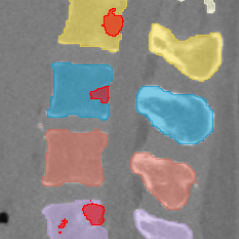}\\
        {\footnotesize (d) Spine}
    \end{minipage}
    \caption{
    Examples of part-aware segmentation across domains:
    (a) natural-image instances with semantic parts from Cityscapes \citep{cordts2016cityscapes}, (b) a BraTS-METS2025 \citep{maleki2025analysis} brain tumor slice showing the metastasis composed of edema (green), non-enhancing (blue), and enhancing (red) sub-regions in brain MR, (c) vertebrae highlighting healthy bone (green), lytic lesions (blue), and sclerotic lesions (red) and (d) distribution of lesions (red) across distinct vertebrae, each represented by a unique color.
    }
    \label{fig:partaware-examples}
\end{figure}

\subsection{Voronoi Matching}\label{sec:voronoi}
The Voronoi variant, introduced as part of the CC-Metrics framework~\citep{jaus2025every}, sidesteps global bipartite matching by partitioning the image domain into per-instance cells: every voxel is assigned to its nearest ground truth segment, and predictions are scored locally inside each cell.
Since each cell contains exactly one ground truth segment, the bipartite construction of \cref{sec:methods} reduces to $|G|$ independent intra-cell matchings, into which any of the previous strategies can be plugged.
The vertex-based \ac{TP}/\ac{FN}/\ac{FP} accounting and \ac{PQ} definition of \cref{eq:tpfnfp,eq:sq} carry over unchanged. 
Predicted segments extending past a cell boundary are re-attributed to the nearest reference rather than counted as an additional \ac{FP}, as depicted in \cref{fig:matching-comparison} (c).
The original CC-Metrics motivation is that local averaging weights every component equally, regardless of size, preventing a few large instances from dominating the score. 
Embedded in our framework, the same construction additionally resolves bipartite matching ambiguity when references are dense or touching.
The formal construction is given in \cref{sec:appendix-voronoi}.

\subsection{Part-Aware Panoptic Matching}\label{sec:part-aware}
Part-aware Panoptic Quality (\textit{PartPQ}), introduced by \citet{degeus2021partaware}, extends \ac{PQ} to hierarchical targets: each scene-level class may carry an internal part decomposition, and a prediction is rewarded only if it both recovers the parent instance \emph{and} agrees with its internal part labeling.
This matches the biomedical setting almost directly: a tumor instance decomposes into edema, non-enhancing, and enhancing sub-regions or a vertebra into body, arch, and process (\cref{fig:partaware-examples}).
PartPQ leaves the matching machinery of \ac{PQ} entirely intact: instance matching is performed exactly as in any of the strategies above on the scene-level class $l$, yielding the same \ac{TP}, \ac{FP}, and \ac{FN} sets via \cref{eq:tpfnfp}.
The only modification is to the \emph{quality term}: the instance-level $\IoU(P_M(g), g)$ in \cref{eq:sq} is replaced by a part-aware quantity $\IoU_p(P_M(g), g)$.
For a class $l$ that carries parts ($l \in \mathcal{L}^{\text{parts}}$), $\IoU_p$ is the mean \ac{IOU} taken over the part labels of $l$ within the matched pair, treating voxels outside both $g$ and $p$ as background. For a class without parts ($l \in \mathcal{L}^{\text{no-parts}}$), it falls back to the instance \ac{IOU}:
\begin{equation}\label{eq:iou-p}
    \IoU_p(P_M(g), g) =
    \begin{cases}
        \mathrm{mean}\, \IoU_{\text{part}}(P_M(g), g), & l \in \mathcal{L}^{\text{parts}} \\
        \IoU_{\text{inst}}(P_M(g), g),                 & l \in \mathcal{L}^{\text{no-parts}}.
    \end{cases}
\end{equation}
For $l \in \mathcal{L}^{\text{parts}}$, the multi-class mean is computed across all part labels of $l$ \emph{plus} the background label assigned to voxels outside the matched segments. Predictions falling on the \emph{void} part label inside $g$ are not counted as part-level \acp{FP} but are still counted as \acp{FN}, mirroring the convention of scene-level \ac{PQ}~\citep{kirillov2019panoptic}.
Substituting $\IoU_p$ into \cref{eq:sq}, the per-class PartPQ is
\begin{equation}\label{eq:partpq}
    \mathrm{PartPQ}_l = \frac{\sum_{g \in \TP} \IoU_p(P_M(g), g)}{|\TP| + \tfrac{1}{2}|\FP| + \tfrac{1}{2}|\FN|},
\end{equation}
which preserves the \ac{SQ}$\times$\ac{RQ} structure of standard \ac{PQ} while embedding the part hierarchy into the quality numerator rather than into a separate matching pass.
Two consequences are worth highlighting.
First, the hierarchy is strict: parts of an unmatched instance ($g \in \FN$ or $p \in \FP$) contribute nothing to the numerator of \cref{eq:partpq}, and are \emph{not} separately re-matched as their own instances.
A necrotic core that is correctly delineated but whose parent tumor was missed at the instance level scores as a single \ac{FN} on the parent, not as a part-level success. 
This is the property that distinguishes PartPQ from the common practice of independently reporting per-region \ac{PQ}~\citep{labella2024analysis}.
Second, PartPQ is orthogonal to the choice of instance matcher: any of the strategies in \cref{sec:one-to-one,sec:many-to-one,sec:one-to-many,sec:voronoi} can supply the parent matching, and $\IoU_p$ is computed downstream from it.
The original formulation~\citep{degeus2021partaware} fixes this matcher to the standard One-to-One rule of \citet{kirillov2019panoptic} with $\tau = 0.5$.
We relax that constraint, so the choice of matcher and the part hierarchy can be configured independently to reflect the failure modes of the underlying data.

%% file: body/panoptica/fig_matching.tex
\begin{figure}[htbp]
    \centering
      \setlength{\fboxrule}{0.5px}
      \fbox{\includegraphics[width=0.98\textwidth]{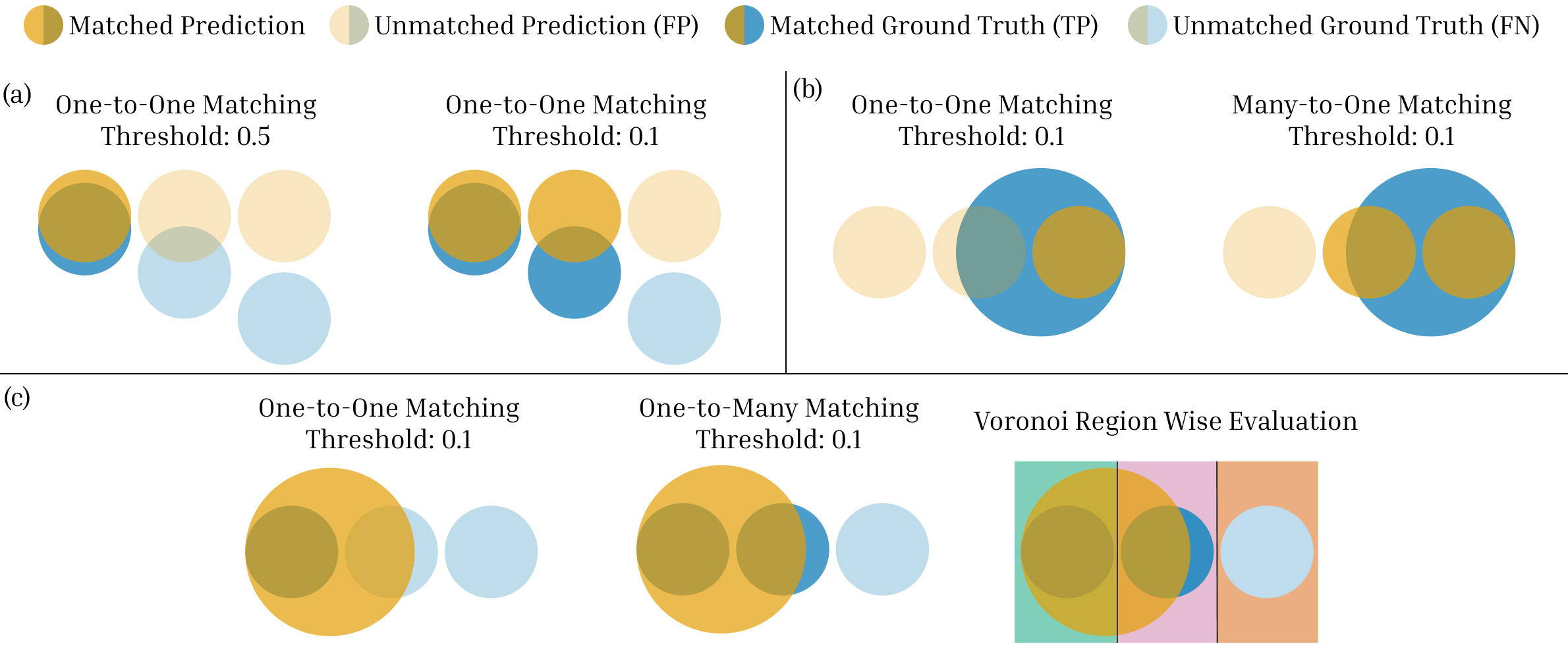}}
    \caption{
    Comparison of different scenarios where the choice of a matching strategy and threshold impacts \ac{PQ} significantly.
    (a) shows a scenario in which at most one predicted segment is matched to a ground truth segment and vice versa. 
       Lowering the threshold counts the center pair as an additional \ac{TP}.
    In scenario (b), two smaller predicted segments overlap a bigger ground truth segment: The Many-to-One matching assigns both overlapping predictions to the ground truth, resulting in a higher score. 
    Scenario (c) shows how One-to-One matching does not credit the partial overlap of the prediction with the middle circle, whereas One-to-Many and Voronoi region-wise evaluation match the prediction with the middle circle as well.
    }
    \label{fig:matching-comparison}
\end{figure}

%% file: body/panoptica/fig_autc.tex
\begin{figure}[htbp]
    \centering
    \includegraphics[width=\linewidth]{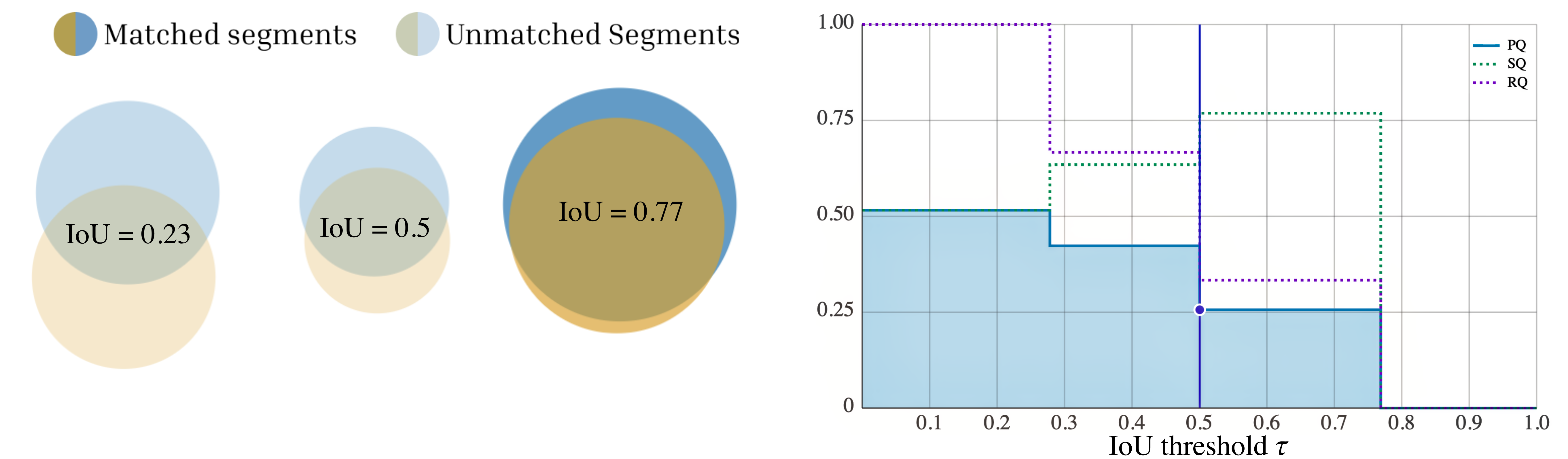}
    \caption{
    Left shows the One-to-One matching for the threshold $\tau = 0.5$.
    The overlapping prediction and ground truth segments in the middle do not exceed the 0.5 threshold and are counted as \ac{FP} and \ac{FN} respectively.
    Right shows the step function mapping all distinct thresholds to their respective values. 
    The \ac{AUTC} reports a summary of the \ac{PQ}, \ac{RQ} and \ac{SQ} over all thresholds.
    Explore this example in our \href{https://koflerlab.org/panoptica-interactive/?refs=0.238\%2C0.401\%2C0.092\%3B0.731\%2C0.421\%2C0.117\%3B0.485\%2C0.416\%2C0.075&preds=0.234\%2C0.542\%2C0.092\%3B0.728\%2C0.456\%2C0.108\%3B0.488\%2C0.481\%2C0.073&matcher=greedy&mode=scene&sap=0&autc=1&asq=0&arq=0}{interactive playground}.
    }
    \label{fig:autc}
\end{figure}

%% file: body/experiments/experiments.tex
\input{body/experiments/coco_table}
\section{Experiments and evaluation tutorials}\label{sec:experiments}
\Cref{tab:coco} shows how different matching strategies behave on the COCO dataset.
Note the asymmetry between Many-to-One and One-to-Many which arises because \acp{TP} are tied to references: One-to-Many scores each of the $k$ references sharing a prediction against the whole prediction, so \ac{SQ} averages $k$ low \acp{IOU}, whereas Many-to-One scores its merged predictions once.
In the appendix, we include perturbation experiments showing how different matching strategies behave when predictions or references are systematically split (\cref{tab:split-predictions} \& \cref{tab:split-references}). 
We publish an \href{https://koflerlab.org/panoptica-interactive}{interactive playground} for exploring thresholds below 0.5 and different matching algorithms. 
The \ac{AUTC} example (\cref{fig:autc}) and toy example (\cref{fig:autc_toy_example}) may be extended and modified or completely new scenarios can be configured.
Further, we provide two Jupyter notebooks showcasing the application of different strategies and part-awareness in the medical domain, including BraTS metastases and part-aware liver lesion segmentation.
All resources are accessible via an \href{https://github.com/koflerlab/Redefining_Instance_Matching-Resources}{umbrella repository}.

%% file: body/experiments/coco_table.tex
\begin{table}[tb]
  \centering
  \setlength{\aboverulesep}{0pt}
  \setlength{\belowrulesep}{0pt}
  \setlength{\tabcolsep}{5pt}
  \renewcommand{\arraystretch}{1.2}
  \caption{\Ac{PQ} on COCO under the three matching strategies at \ac{IOU} threshold $\tau = 0.1$, split into thing (Th) and stuff (St) classes \cite{linMicrosoftCOCOCommon2014, kirillovPanopticFeaturePyramid2019, chengMaskedattentionMaskTransformer2022, NEURIPS2021_950a4152}. 
  All values are COCO-style means over categories, $\times 100$. 
  Stuff values do not depend on the matching strategy, because each stuff category occurs at most once per image in both ground truth and prediction.}
  \label{tab:coco}
  \begin{tabular*}{\linewidth}{@{\extracolsep{\fill}}ll|c|ccc|ccc@{}}
    \toprule
    Model & Matching & PQ
      & PQ$^\mathrm{Th}$ & SQ$^\mathrm{Th}$ & RQ$^\mathrm{Th}$
      & PQ$^\mathrm{St}$ & SQ$^\mathrm{St}$ & RQ$^\mathrm{St}$ \\
    \midrule
    mask2former-swin-large & 1:1 & 58.97 & 64.78 & 80.88 & 79.59 & 50.20 & 74.46 & 66.28 \\
                             & M:1 & 59.75 & 66.07 & 81.57 & 80.60 & 50.20 & 74.46 & 66.28 \\
                             & 1:M & 58.89 & 64.66 & 79.33 & 81.01 & 50.20 & 74.46 & 66.28 \\
    \midrule
    mask2former-swin-base  & 1:1 & 56.54 & 61.91 & 79.71 & 77.17 & 48.43 & 74.07 & 64.16 \\
                             & M:1 & 57.36 & 63.27 & 80.49 & 78.18 & 48.43 & 74.07 & 64.16 \\
                             & 1:M & 56.46 & 61.79 & 78.03 & 78.68 & 48.43 & 74.07 & 64.16 \\
    \midrule
    maskformer-swin-base   & 1:1 & 53.61 & 58.53 & 76.58 & 75.67 & 46.19 & 70.22 & 64.17 \\
                             & M:1 & 54.67 & 60.29 & 77.59 & 77.03 & 46.19 & 70.22 & 64.17 \\
                             & 1:M & 53.55 & 58.43 & 74.92 & 77.26 & 46.19 & 70.22 & 64.17 \\
    \midrule
    panoptic-fpn-R-50-3x & 1:1 & 44.37 & 50.29 & 78.02 & 63.46 & 35.45 & 61.59 & 53.16 \\
                             & M:1 & 44.82 & 51.04 & 78.61 & 63.99 & 35.45 & 61.59 & 53.16 \\
                             & 1:M & 44.36 & 50.26 & 75.95 & 65.12 & 35.45 & 61.59 & 53.16 \\
    \bottomrule
  \end{tabular*}
\end{table}

%% file: body/discussion/discussion.tex
\section{Discussion}\label{sec:discussion}

We recast \ac{PQ} as a constrained bipartite assignment, anchor \ac{TP}/\ac{FP}/\ac{FN} counts to vertices rather than edges, preserving $|\TP|+|\FN|=|G|$ and introduce a taxonomy of matching strategies by constraining degrees of vertices.
On top of this, we build a threshold-agnostic summary (\ac{AUTC}) and a part-aware extension that decouples PartPQ from the original \ac{PQ} constraints, so matching and part hierarchy can be configured independently.
The vertex-based accounting is what lets these three strategies share a metric in the first place.
Under any asymmetric matching the edge count $|M|$ can exceed $|G|$ or $|P|$.
Identifying \acp{TP} with edges would push \ac{RQ} above $1$ and dissolve its interpretation as a recognition rate.
Anchoring \ac{TP} and \ac{FN} to ground-truth vertices closes this gap, so \ac{RQ} stays directly comparable across One-to-One, Many-to-One, and One-to-Many matchings.
Across the toy example and case studies, the choice of strategy maps onto a tension between \ac{RQ} and \ac{SQ}.
One-to-One enforces a strict per-object assignment and is the only strategy that penalizes \ac{OS} and \ac{US} symmetrically.
The max-weight greedy variant occasionally trades a marginal recognition for a higher accumulated overlap when two predictions cover the same reference, but in practice the two coincide in the vast majority of cases.
Many-to-One greedy merge is the most lenient strategy when predictions are split, yet it still requires a single edge above $\tau$ to register a detection: in the toy example, this produces a sharp drop above $\tau \approx 0.2$ even when accumulated overlap across the reference remains high, so a heavily fragmented prediction can still be counted as \iac{FN}.
Reporting \ac{AUTC} bypasses the dependence on a single threshold.
\input{body/panoptica/fig_autc_toy_example}
\paragraph{Practitioner guidelines.}
In many segmentation datasets, instance labels do not exist and must be approximated from semantic labels, for example, via connected component analysis. 
In extreme cases, this can merge two components that are only connected by a single pixel but represent individual instances. 
Models that correctly predict the individual instances get punished by a One-to-One matching. 
In these cases, the leniency of Many-to-One matching towards \ac{OS} can be an appropriate choice.
If instance labels exist and expert disagreement is negligible, One-to-One matching is the idiomatic choice.
If reference instances are split by partial occlusion and recovered as fragmented instances or merging references  is acceptable in the downstream application, One-to-Many matching is an appropriate choice.
However, we emphasize that in many scenarios, the evaluation with a combination of multiple metrics is beneficial. 
Having a variety of metrics can serve as an analytical tool, making model training and comparisons across models more interpretable by revealing insights into model performance across failure modes. 
On the other hand, omitting specific metrics may also hide specific failure modes that are only obvious in certain configurations.
\\
The interaction between matcher and part hierarchy is most consequential under Many-to-One.
When several adjacent fragments are credited as a single parent instance, their union defines the part-level $\IoU_p$, so a fragmented prediction whose parts are individually correct can recover full part-level credit, whereas the original One-to-One rule of \citet{degeus2021partaware} would reject the whole instance and discard its part labeling along with it.

\noindent\textbf{Limitations, future work, and conclusion.}\label{sec:limitations}
Many-to-Many matching is intentionally outside our framework because the bipartite formulation breaks down and lacks interpretability once both degree constraints are dropped.
Related approaches such as Cluster Dice~\citep{kundu2025cluster} address the same failure mode through clustering, and a unified treatment that bridges assignment and clustering views remains open.
Many-to-One and One-to-Many expose an asymmetry which follows from the reference centric definition of \ac{TP}/\ac{FN} and the adjusted \ac{SQ} definition which counts each segment once for Many-to-One but may double count predictions when references are merged for One-to-Many.
The suggested greedy merge for Many-to-One is based on heuristics without an established optimality guarantee against $M^\star_{\text{PQ}}$ (\cref{sec:optimal-matching}). 
Tighter algorithmic bounds or an exact tractable reformulation would put the strategy on firmer ground.
Our empirical study is limited in scope: we evaluate four pretrained COCO models at a single threshold and under controlled perturbations, threshold sweeps are shown only on toy examples, and PartPQ is illustrated on biomedical examples without a quantitative study.

We present a unified framework for \ac{PQ} below the standard $IoU = 0.5$ threshold, recasting segment matching as a constrained bipartite assignment in which vertex-based accounting keeps \ac{PQ} consistent across One-to-One, Many-to-One, One-to-Many, and Voronoi matchings.
We release the presented strategies as an extension to the unified open-source package Panoptica, so practitioners can configure the matcher, threshold, and part hierarchy to match their specific requirements.

%% file: body/panoptica/fig_autc_toy_example.tex
\begin{figure}[tb]
    \centering
    \includegraphics[width=0.34\linewidth]{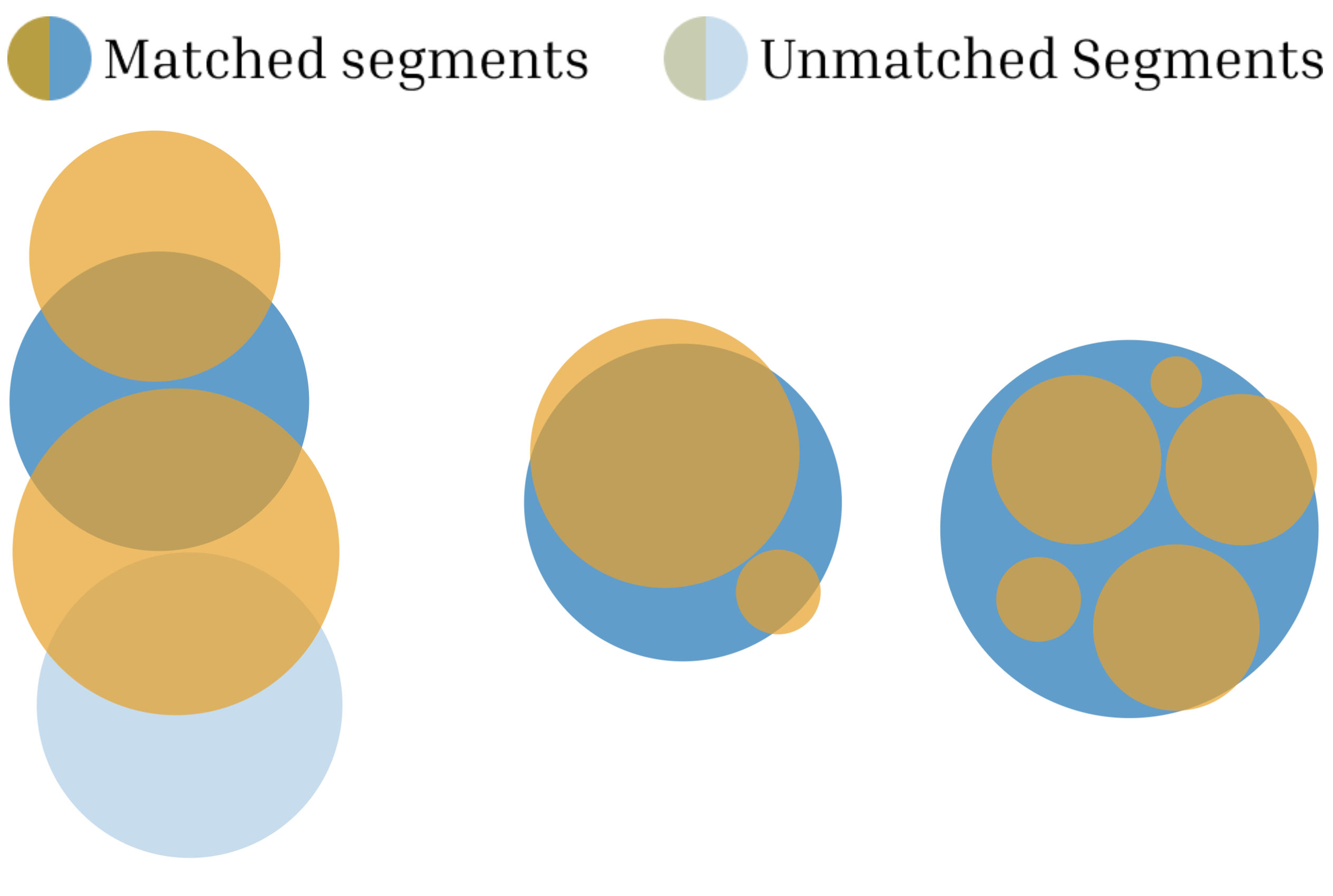}
    \includegraphics[width=0.65\linewidth]{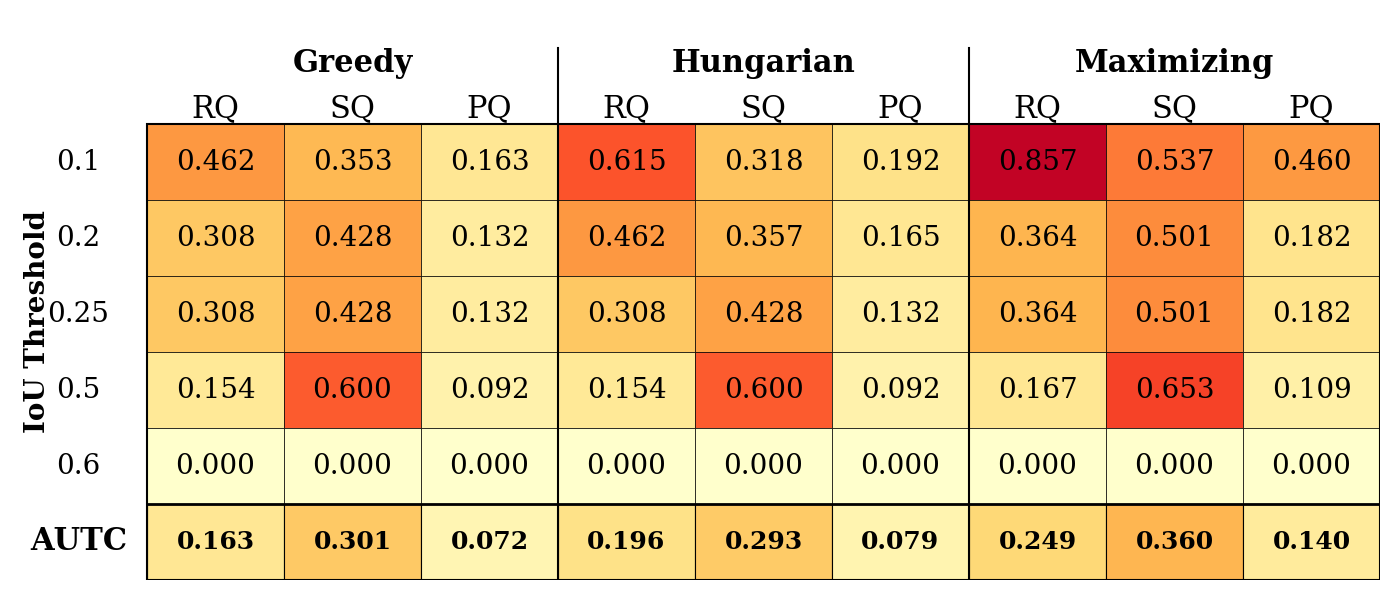}
    \caption{
    A toy example showcasing different scenarios in real-life segmentation tasks, like fragmented segmentation of a larger object, and the corresponding \ac{SQ}, \ac{RQ}, and \ac{PQ} values when setting different thresholds.
    The left example shows a Many-to-One matching for the threshold $\tau = 0.2$.
    The choice of a matching strategy heavily influences all resulting metric computations as demonstrated in the table on the right. 
    The color saturation of cells encodes the reported quality.
    Adjust the matching strategy or the samples in our \href{https://koflerlab.org/panoptica-interactive/?refs=0.769,0.482,0.125;0.474,0.453,0.105;0.128,0.342,0.099;0.148,0.675,0.101\&preds=0.734,0.406,0.056;0.8,0.59,0.055;0.843,0.417,0.05;0.8,0.321,0.017;0.709,0.559,0.028;0.537,0.551,0.028;0.462,0.399,0.089;0.125,0.183,0.083;0.139,0.507,0.108\&matcher=maximize-merge\&mode=scene\&sap=0\&autc=1\&asq=1\&arq=1}{interactive playground}.
    }
    \label{fig:autc_toy_example}
\end{figure}

%% file: body/acknowledgements/acknowledgements.tex
\begin{ack}
This work was supported by Anna Valentina Lioba Eleonora Claire Javid Mamasani and the Gemeinnützige Hertie-Stiftung.
SSK acknowledges support from a grant from the United Kingdom Research and Innovation Medical Research Council awarded as part of a Doctoral Training Partnership at King's College London (MR/W006820/1).
TV acknowledges support from Medtronic/RAEng [RCSRF1819\textbackslash{}7\textbackslash{}34] and Wellcome/EPSRC [WT203148/Z/16/Z; NS/A000049/1].
The authors thank the International Max Planck Research School for Intelligent Systems (IMPRS-IS) for supporting EG and NM.
The authors declare no competing interests.
\end{ack}

%% file: body/appendix/appendix.tex
\input{body/appendix/perturbations}
\input{body/appendix/voronoi}
\input{body/appendix/optimal_matching}
\input{body/appendix/RAM}

\section{Figures} \label{sec:appendix_figures}
\input{body/preliminaries/figures/fig_segmentation_types}

%% file: body/appendix/perturbations.tex
\begin{table*}[h]
  \centering
  \setlength{\tabcolsep}{4pt}
  \caption{COCO centroid splitting of \emph{predicted} segments (Things).
    PQ, SQ and RQ in \% per matching strategy with threshold $\tau=0.1$. $p$ is the split probability.
    M:1 is lenient towards oversegmentation. Fragmented predictions are not penalized as FP and aggregated in the SQ IoU computation.
    For $p>0$: mean $\pm$ sample std over 5 random split draws with fixed model weights.}
  \label{tab:split-predictions}
  \begin{tabular*}{\textwidth}{@{\extracolsep{\fill}}l ccc ccc ccc@{}}
    \toprule
        & \multicolumn{3}{c}{1:1} & \multicolumn{3}{c}{M:1} & \multicolumn{3}{c}{1:M} \\
    \cmidrule(lr){2-4} \cmidrule(lr){5-7} \cmidrule(l){8-10}
    $p$ & PQ & SQ & RQ & PQ & SQ & RQ & PQ & SQ & RQ \\
    \midrule
    \multicolumn{10}{@{}l}{\textit{mask2former-swin-large}} \\
    0 & 64.8 & 80.9 & 79.6 & 66.1 & 81.6 & 80.6 & 64.7 & 79.3 & 81.0 \\
    0.5 & 41.6{\scriptsize$\pm$0.3} & 63.7{\scriptsize$\pm$0.4} & 64.9{\scriptsize$\pm$0.2} & 63.8{\scriptsize$\pm$0.0} & 81.6{\scriptsize$\pm$0.0} & 77.7{\scriptsize$\pm$0.0} & 41.6{\scriptsize$\pm$0.3} & 63.1{\scriptsize$\pm$0.4} & 65.7{\scriptsize$\pm$0.2} \\
    \addlinespace
    \multicolumn{10}{@{}l}{\textit{mask2former-swin-base}} \\
    0 & 61.9 & 79.7 & 77.2 & 63.3 & 80.5 & 78.2 & 61.8 & 78.0 & 78.7 \\
    0.5 & 40.1{\scriptsize$\pm$0.2} & 63.0{\scriptsize$\pm$0.2} & 63.3{\scriptsize$\pm$0.1} & 61.0{\scriptsize$\pm$0.1} & 80.5{\scriptsize$\pm$0.1} & 75.3{\scriptsize$\pm$0.1} & 40.2{\scriptsize$\pm$0.2} & 62.3{\scriptsize$\pm$0.2} & 64.1{\scriptsize$\pm$0.1} \\
    \addlinespace
    \multicolumn{10}{@{}l}{\textit{maskformer-swin-base}} \\
    0 & 58.5 & 76.6 & 75.7 & 60.3 & 77.6 & 77.0 & 58.4 & 74.9 & 77.3 \\
    0.5 & 37.9{\scriptsize$\pm$0.1} & 60.9{\scriptsize$\pm$0.1} & 61.6{\scriptsize$\pm$0.1} & 57.4{\scriptsize$\pm$0.1} & 77.6{\scriptsize$\pm$0.0} & 73.2{\scriptsize$\pm$0.1} & 37.9{\scriptsize$\pm$0.1} & 60.2{\scriptsize$\pm$0.1} & 62.5{\scriptsize$\pm$0.1} \\
    \addlinespace
    \multicolumn{10}{@{}l}{\textit{panoptic-fpn-R-50-3x}} \\
    0 & 50.3 & 78.0 & 63.5 & 51.0 & 78.6 & 64.0 & 50.3 & 76.0 & 65.1 \\
    0.5 & 33.1{\scriptsize$\pm$0.2} & 60.9{\scriptsize$\pm$0.2} & 53.6{\scriptsize$\pm$0.1} & 49.4{\scriptsize$\pm$0.1} & 78.4{\scriptsize$\pm$0.1} & 62.1{\scriptsize$\pm$0.1} & 33.2{\scriptsize$\pm$0.2} & 60.0{\scriptsize$\pm$0.2} & 54.5{\scriptsize$\pm$0.1} \\
    \bottomrule
  \end{tabular*}
  \bigskip
  \caption{COCO centroid splitting of \emph{reference} segments (Things).
    PQ, SQ and RQ in \% per matching strategy with threshold $\tau=0.1$. $p$ is the split probability.
    1:M RQ is lenient towards undersegmentation but its SQ term gets diluted by low IoU from single predictions shared by many references.
    For $p>0$: mean $\pm$ sample std over 5 random split draws with fixed model weights.}
  \label{tab:split-references}
  \begin{tabular*}{\textwidth}{@{\extracolsep{\fill}}l ccc ccc ccc@{}}
    \toprule
        & \multicolumn{3}{c}{1:1} & \multicolumn{3}{c}{M:1} & \multicolumn{3}{c}{1:M} \\
    \cmidrule(lr){2-4} \cmidrule(lr){5-7} \cmidrule(l){8-10}
    $p$ & PQ & SQ & RQ & PQ & SQ & RQ & PQ & SQ & RQ \\
    \midrule
    \multicolumn{10}{@{}l}{\textit{mask2former-swin-large}} \\
    0 & 64.8 & 80.9 & 79.6 & 66.1 & 81.6 & 80.6 & 64.7 & 79.3 & 81.0 \\
    0.5 & 41.0{\scriptsize$\pm$0.3} & 63.5{\scriptsize$\pm$0.3} & 64.2{\scriptsize$\pm$0.2} & 41.6{\scriptsize$\pm$0.3} & 64.0{\scriptsize$\pm$0.3} & 64.6{\scriptsize$\pm$0.2} & 45.9{\scriptsize$\pm$0.2} & 55.1{\scriptsize$\pm$0.3} & 83.0{\scriptsize$\pm$0.0} \\
    \addlinespace
    \multicolumn{10}{@{}l}{\textit{mask2former-swin-base}} \\
    0 & 61.9 & 79.7 & 77.2 & 63.3 & 80.5 & 78.2 & 61.8 & 78.0 & 78.7 \\
    0.5 & 39.3{\scriptsize$\pm$0.4} & 62.8{\scriptsize$\pm$0.4} & 62.2{\scriptsize$\pm$0.2} & 39.9{\scriptsize$\pm$0.4} & 63.4{\scriptsize$\pm$0.4} & 62.6{\scriptsize$\pm$0.2} & 44.1{\scriptsize$\pm$0.2} & 54.5{\scriptsize$\pm$0.3} & 80.6{\scriptsize$\pm$0.1} \\
    \addlinespace
    \multicolumn{10}{@{}l}{\textit{maskformer-swin-base}} \\
    0 & 58.5 & 76.6 & 75.7 & 60.3 & 77.6 & 77.0 & 58.4 & 74.9 & 77.3 \\
    0.5 & 37.9{\scriptsize$\pm$0.3} & 60.7{\scriptsize$\pm$0.3} & 61.8{\scriptsize$\pm$0.2} & 38.6{\scriptsize$\pm$0.3} & 61.4{\scriptsize$\pm$0.3} & 62.3{\scriptsize$\pm$0.2} & 42.4{\scriptsize$\pm$0.2} & 52.8{\scriptsize$\pm$0.3} & 79.7{\scriptsize$\pm$0.1} \\
    \addlinespace
    \multicolumn{10}{@{}l}{\textit{panoptic-fpn-R-50-3x}} \\
    0 & 50.3 & 78.0 & 63.5 & 51.0 & 78.6 & 64.0 & 50.3 & 76.0 & 65.1 \\
    0.5 & 31.5{\scriptsize$\pm$0.2} & 62.4{\scriptsize$\pm$0.4} & 49.8{\scriptsize$\pm$0.1} & 31.8{\scriptsize$\pm$0.2} & 62.8{\scriptsize$\pm$0.3} & 50.0{\scriptsize$\pm$0.1} & 35.9{\scriptsize$\pm$0.1} & 53.8{\scriptsize$\pm$0.4} & 66.1{\scriptsize$\pm$0.1} \\
    \bottomrule
  \end{tabular*}
\end{table*}

%% file: body/appendix/voronoi.tex
\section{Voronoi Matching: Formal Construction}\label{sec:appendix-voronoi}

Let $\Omega$ denote the image domain and $d(x, g)$ the (Euclidean) distance from voxel $x \in \Omega$ to the nearest voxel of ground truth segment $g \in G$.
The Voronoi cell of $g$ is
\begin{equation}\label{eq:voronoi-cell}
    V(g) = \{ x \in \Omega : d(x, g) \leq d(x, g') \;\; \forall g' \in G \},
\end{equation}
with ties broken arbitrarily so that $\{V(g)\}_{g \in G}$ partitions $\Omega$.
For each ground truth segment we define its \emph{local prediction} as the restriction of all predicted segments to its cell,
\begin{equation}\label{eq:voronoi-local-pred}
    \tilde p(g) = \Big( \bigcup_{p \in P} p \Big) \cap V(g),
\end{equation}
and replace the bipartite edge set of \cref{eq:edges} with the per-cell edges
\begin{equation}\label{eq:voronoi-edges}
    E^{\text{Vor}}_\tau = \{ (g, \tilde p(g)) : g \in G,\; \IoU(g, \tilde p(g)) > \tau \}.
\end{equation}
The resulting matching $M^{\text{Vor}} = E^{\text{Vor}}_\tau$ is one-to-one by construction---each ground truth segment is paired with the unique aggregated prediction inside its cell---so the \ac{TP}/\ac{FN}/\ac{FP} accounting of \cref{eq:tpfnfp} and the \ac{SQ}/\ac{RQ}/\ac{PQ} definitions of \cref{eq:sq} apply unchanged.
Predicted mass that falls outside every cell never contributes to \iac{TP} but is still penalised through the global \ac{FP} count.

The construction generalises directly: rather than aggregating all predictions inside a cell into a single $\tilde p(g)$, the chosen instance matcher (One-to-One, Many-to-One, or One-to-Many) can be applied independently within each cell, and the per-cell results are aggregated into the global \ac{TP}/\ac{FN}/\ac{FP} counts.
This is the form realised by Panoptica's region-wise pipeline, in which a Voronoi partition is computed once on the reference and the standard bipartite matching is run inside every cell.

%% file: body/appendix/optimal_matching.tex
\section{Optimal Matching Across Strategies} \label{sec:optimal-matching}

A matching strategy constrains the structure of $M$ but does not pin it down: multiple matchings may satisfy a strategy's degree constraints.
We must therefore specify an objective that selects an optimal matching $M^\star$.
For a strategy $S$, let $\mathcal{M}_S(\tau)$ denote the matchings $M \subseteq E_\tau$ satisfying its degree constraints, and note that, since \cref{eq:tpfnfp} defines \acp{FP} as the unmatched predicted segments, $P \setminus \FP$ is the set of matched ones.

With the counts of \cref{eq:tpfnfp},
\begin{equation}\label{eq:pq-matching}
    \mathrm{PQ}(M) = \frac{\sum_{g \in \TP} \IoU(P_M(g), g)}
                          {\tfrac{1}{2}(|G| + |P|) + \tfrac{1}{2}(|\TP| - |P \setminus \FP|)}.
\end{equation}
The denominator is constant under One-to-One ($|\TP| = |P \setminus \FP|$), grows when One-to-Many reuses a predicted segment, and shrinks when Many-to-One absorbs fragments.
For $\tau \geq 0.5$, every edge in $E_\tau$ has $\IoU > 0.5$ strictly, and since segments do not overlap within $G$ or within $P$, \citet[Theorem~1]{kirillov2019panoptic} bounds every vertex degree in $E_\tau$ by one.
$E_\tau$ is then itself a One-to-One matching, every degree-constrained strategy returns $M^\star = E_\tau$, and the objective matters only for $\tau < 0.5$.

Since \ac{PQ} is the metric we ultimately report, a natural reference objective is to optimise it directly,
\begin{equation}\label{eq:max-pq}
    M^\star_{\text{PQ}} = \argmax_{M \in \mathcal{M}_S(\tau)} \mathrm{PQ}(M).
\end{equation}
\ac{PQ} couples edges through the union $P_M(g)$ in the numerator and through $|\TP|$ and $|\FP|$ in the denominator of \cref{eq:pq-matching}, so it does not decompose into a sum over edges and admits no obvious polynomial algorithm in general.

We therefore work with an edge-decomposable surrogate, the \emph{max-weight matching}
\begin{equation}\label{eq:max-weight}
    M^\star_{\text{weight}} = \argmax_{M \in \mathcal{M}_S(\tau)} \sum_{(g,p) \in M} \IoU(g, p),
\end{equation}
and ask, for each strategy, whether $M^\star_{\text{weight}}$ coincides with $M^\star_{\text{PQ}}$, and if not, which matching we adopt as the operational definition.
Ties between equally weighted matchings are broken deterministically.

\paragraph{One-to-One.}
Under the one-to-one constraint, $M^\star_{\text{weight}}$ and $M^\star_{\text{PQ}}$ coincide.
Every matched ground truth has $|P_M(g)| = 1$, so the numerator of \cref{eq:pq-matching} equals the edge-weight sum, and $|\TP| = |P \setminus \FP| = |M|$ makes the denominator $\tfrac{1}{2}(|G| + |P|)$ independent of $M$.
Hence \ac{PQ} is monotone in the edge-weight sum, and
\begin{equation}\label{eq:1to1-opt}
    M^\star = \argmax_{M \in \mathcal{M}_{\text{1:1}}(\tau)} \sum_{(g,p) \in M} \IoU(g, p)
\end{equation}
solves \eqref{eq:max-pq} exactly.
$M^\star$ is computable in $O((m+n)^3)$ time by the Hungarian algorithm~\citep{kuhn1955hungarian}.
For $\tau < 0.5$, multiple ground truth segments can compete for the same predicted segment and vice versa, so a global assignment is required.
At $\tau = 0.5$, our formulation reduces to the original \ac{PQ} matching.

\paragraph{Many-to-One.}
Here the surrogate diverges from max-\ac{PQ}, because it credits each predicted segment with its own $\IoU(g, p)$ although the segments matched to $g$ are scored once, through $\IoU(P_M(g), g)$.
Absorbing an additional predicted segment into $g$ removes one \ac{FP} from the denominator of \cref{eq:pq-matching} and raises \ac{PQ} exactly when the change in $\IoU(P_M(g), g)$ exceeds $-\mathrm{PQ}(M)/2$.
Since every edge adds positive weight, $M^\star_{\text{weight}}$ can also absorb segments that dilute $\IoU(P_M(g), g)$ by more.
We therefore drop the proxy and target the \ac{SQ} contribution per ground truth directly.
We are not aware of an exact polynomial algorithm for $M^\star_{\text{PQ}}$ under this constraint and perform a greedy merge restricted to edges in $E_\tau$, as the operational definition: it admits a candidate predicted segment only when it strictly improves $\IoU(P_M(g), g)$.
This avoids the dilution failure mode, but is a heuristic with its own objective rather than an approximation of $M^\star_{\text{PQ}}$, which may itself favour absorbing segments that slightly lower $\IoU(P_M(g), g)$.

\paragraph{One-to-Many.}
Because $\deg_M(g) \leq 1$, each $g \in \TP$ has $|P_M(g)| = 1$, and the numerator of \cref{eq:pq-matching} again equals the edge-weight sum.
Moreover, ground truth segments do not compete for predicted segments: a single predicted segment may match arbitrarily many ground truth segments.
The max-weight matching therefore reduces to selecting, independently for each ground truth segment with at least one incident edge, its highest-\ac{IOU} predicted segment:
\begin{equation}\label{eq:1tom-opt}
    M^\star = \bigl\{\bigl(g,\, \argmax_{p : (g,p) \in E_\tau} \IoU(g,p)\bigr) : g \in G,\ \deg_{E_\tau}(g) \geq 1\bigr\},
\end{equation}
which is computable in $O(mn)$ time.
Unlike in the One-to-One case, $M^\star$ need not maximise \ac{PQ}: assigning $g$ to an already matched predicted segment rather than to an otherwise unmatched one can increase the edge-weight sum while leaving the latter as \iac{FP}, which enlarges the denominator of \cref{eq:pq-matching}.
We nonetheless adopt max-weight as the operational definition, because it credits each ground truth segment with its best-overlapping predicted segment independently of all other ground truth segments.

\paragraph{Voronoi.}
The partition $\{V(g)\}$ of \cref{sec:appendix-voronoi} depends only on the ground truth, not on $P$, and decomposes the matching into $|G|$ independent subproblems with a single ground truth segment each.
Within a cell, One-to-One and One-to-Many coincide and select the highest-\ac{IOU} predicted segment, while Many-to-One applies the greedy merge above to the predicted segments restricted to that cell.
Constructing the cells reduces to a single nearest-label distance transform on $G$, computable in $O(|\Omega|)$ time, with ties between equidistant ground truth segments broken deterministically.
Each per-cell \ac{IOU} is then evaluated in time linear in $|V(g)|$.

\paragraph{Part-Aware.}
We do not optimise PartPQ directly.
The parent matching $M^\star$ is computed on the instance-level \ac{IOU} by the optimiser of the chosen instance strategy (\cref{eq:1to1-opt,eq:1tom-opt}, the greedy merge above for Many-to-One, or the Voronoi partition of \cref{sec:voronoi}). 
$\IoU_p$ is then evaluated per matched pair in time linear in the matched volume.
In particular, the exactness of the One-to-One case does not carry over: maximising instance-level \ac{IOU} need not maximise PartPQ.
Optimising the parent matching for $\mathrm{PartPQ}$ rather than for the scene-level \ac{PQ} would couple part- and instance-level overlaps and is, to our knowledge, not addressed in the literature.
We treat the parent-then-parts decomposition as the operational definition.

%% file: body/appendix/RAM.tex
\section{Compute Resources}\label{sec:compute}
All experiments run on CPU.
Table~\ref{tab:ram} reports peak RAM for the notebooks, with all intermediate steps tracked (Full) and with tracking disabled (Stripped), which suffices to reproduce the reported results.
The vertebral example requires the most memory due to its large volume and region-wise evaluation.
The COCO experiment peaks at 2.1\,GB.

\begin{table}[h]
\centering
\caption{Peak RAM usage (GB) of the example notebooks.}\label{tab:ram}
\begin{tabular}{llrr}
\hline
Example & Shape & Full & Stripped \\
\hline
Vertebral & $512{\times}512{\times}1031$ & 32 & 21 \\
Liver & $512{\times}367{\times}422$ & 9 & 6 \\
Brain-Mets & $240{\times}240{\times}155$ & 5 & 2.5 \\
\hline
\end{tabular}
\end{table}

%% file: body/preliminaries/figures/fig_segmentation_types.tex
\begin{figure}[htbp]
    \centering
    \includegraphics[width=1\linewidth]{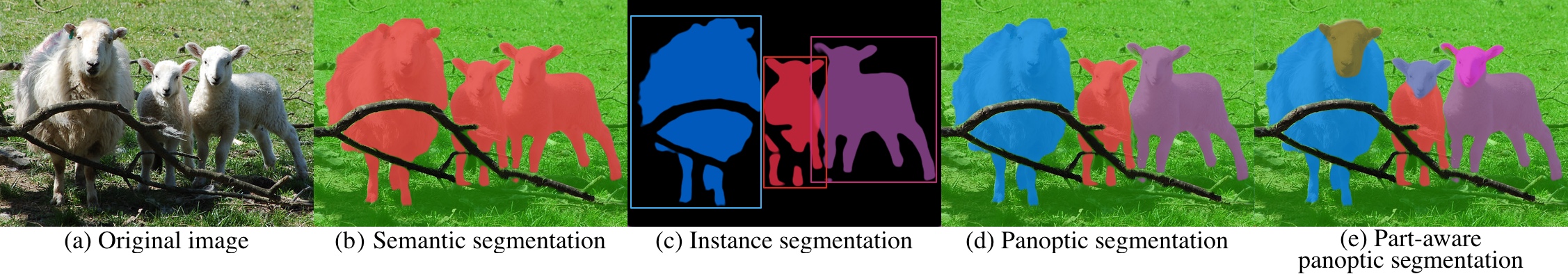}
    \vspace{-15pt}
    \caption{
    Semantic segmentation delineates \textit{stuff} classes exclusively (b), whereas panoptic segmentation augments semantic classes with \textit{things} which are countable objects like individual sheep (d)\cite{linMicrosoftCOCOCommon2014}. 
    Instance segmentation permits multiple instance assignments per pixel in contrast to panoptic segmentation (c).
    Part-aware panoptic segmentation combines scene and part parsing in a single task (e).
    }
    \label{fig:tasks}
\end{figure}

%% file: checklist.tex
\section*{NeurIPS Paper Checklist}

\begin{enumerate}

\item {\bf Claims}
    \item[] Question: Do the main claims made in the abstract and introduction accurately reflect the paper's contributions and scope?
    \item[] Answer: \answerYes{}
    \item[] Justification: The introduction clearly outlines the unified framework for part-aware matching in panoptic segmentation, focusing on the medical imaging domain and the implementation of three matching strategies (One-to-One, Many-to-One, One-to-Many) and AUTC.

\item {\bf Limitations}
    \item[] Question: Does the paper discuss the limitations of the work performed by the authors?
    \item[] Answer: \answerYes{} %
    \item[] Justification: Please refer to \Cref{sec:limitations}.

\item {\bf Theory assumptions and proofs}
    \item[] Question: For each theoretical result, does the paper provide the full set of assumptions and a complete (and correct) proof?
    \item[] Answer: \answerNA{}
    \item[] Justification: The paper focuses on different matching strategies and empirical evaluation rather than formal mathematical proofs or new theoretical theorems. A discussion about optimality is included in the appendix. \cref{sec:optimal-matching}.

    \item {\bf Experimental result reproducibility}
    \item[] Question: Does the paper fully disclose all the information needed to reproduce the main experimental results of the paper to the extent that it affects the main claims and/or conclusions of the paper (regardless of whether the code and data are provided or not)?
    \item[] Answer: \answerYes{}
    \item[] Justification: We provide open-source software framework and notebooks to reproduce all results. (\url{https://github.com/koflerlab/Redefining_Instance_Matching-Resources})

\item {\bf Open access to data and code}
    \item[] Question: Does the paper provide open access to the data and code, with sufficient instructions to faithfully reproduce the main experimental results, as described in supplemental material?
    \item[] Answer: \answerYes{}
    \item[] Justification: Yes, our repository (\url{https://github.com/koflerlab/Redefining_Instance_Matching-Resources}) provides links to all interactive tutorials.
    Furthermore, our toy examples are available for reproduction of results.

\item {\bf Experimental setting/details}
    \item[] Question: Does the paper specify all the training and test details (e.g., data splits, hyperparameters, how they were chosen, type of optimizer) necessary to understand the results?
    \item[] Answer: \answerYes{}
    \item[] Justification: We did not train models for this paper. The case studies in \Cref{sec:experiments} use published predictions from BraTS-METS2025 and liver-lesion benchmarks, the COCO experiment uses standard pretrained models and the dataset references and evaluation protocols are documented in our repository (\url{https://github.com/koflerlab/Redefining_Instance_Matching-Resources}).

\item {\bf Experiment statistical significance}
    \item[] Question: Does the paper report error bars suitably and correctly defined or other appropriate information about the statistical significance of the experiments?
    \item[] Answer: \answerYes{}
    \item[] Justification:
    Results with random segment splitting are reported as mean ± sample std over 5 independent split draws with fixed model weights. All other evaluations are deterministic given the released checkpoints, so repeated runs give identical results.

\item {\bf Experiments compute resources}
    \item[] Question: For each experiment, does the paper provide sufficient information on the computer resources (type of compute workers, memory, time of execution) needed to reproduce the experiments?
    \item[] Answer: \answerYes{}
    \item[] Justification: Appendix~\ref{sec:compute} reports the hardware and peak RAM per experiment.
    
\item {\bf Code of ethics}
    \item[] Question: Does the research conducted in the paper conform, in every respect, with the NeurIPS Code of Ethics \url{https://neurips.cc/public/EthicsGuidelines}?
    \item[] Answer: \answerYes{}
    \item[] Justification:
    All methodologies are transparently documented through an open-source framework to ensure reproducibility, while our toy examples utilize simple geometric shapes to maintain strict privacy.
    Our work focuses on improving segmentation evaluation in biomedical imaging, posing no risks to human rights, safety, or the environment.

\item {\bf Broader impacts}
    \item[] Question: Does the paper discuss both potential positive societal impacts and negative societal impacts of the work performed?
    \item[] Answer: \answerNA{} %
    \item[] Justification:
    The paper is foundational research and is not tied to any particular application.
    Neither direct positive, nor negative societal implications are to be expected. 
    
\item {\bf Safeguards}
    \item[] Question: Does the paper describe safeguards that have been put in place for responsible release of data or models that have a high risk for misuse (e.g., pre-trained language models, image generators, or scraped datasets)?
    \item[] Answer: \answerNA{}
    \item[] Justification:
    We neither train, nor release models.
    The toy examples we release are simple geometric shapes not posing any risk to animals nor humans.

\item {\bf Licenses for existing assets}
    \item[] Question: Are the creators or original owners of assets (e.g., code, data, models), used in the paper, properly credited and are the license and terms of use explicitly mentioned and properly respected?
    \item[] Answer: \answerYes{}
    \item[] Justification: Yes, we properly cite and credit all datasets we use in both the manuscript and released tutorials.

\item {\bf New assets}
    \item[] Question: Are new assets introduced in the paper well documented and is the documentation provided alongside the assets?
    \item[] Answer: \answerYes{}
    \item[] Justification: We release the toy datasets of geometric shapes and the open-source matching framework via the umbrella repository (\url{https://github.com/koflerlab/Redefining_Instance_Matching-Resources}), with documentation provided alongside.

\item {\bf Crowdsourcing and research with human subjects}
    \item[] Question: For crowdsourcing experiments and research with human subjects, does the paper include the full text of instructions given to participants and screenshots, if applicable, as well as details about compensation (if any)? 
    \item[] Answer: \answerNA{}
    \item[] Justification: Neither crowdsourcing, nor research with human subjects was carried out.

\item {\bf Institutional review board (IRB) approvals or equivalent for research with human subjects}
    \item[] Question: Does the paper describe potential risks incurred by study participants, whether such risks were disclosed to the subjects, and whether Institutional Review Board (IRB) approvals (or an equivalent approval/review based on the requirements of your country or institution) were obtained?
    \item[] Answer: \answerNA{}
    \item[] Justification: The paper does not involve crowdsourcing nor research with human subjects.

\item {\bf Declaration of LLM usage}
    \item[] Question: Does the paper describe the usage of LLMs if it is an important, original, or non-standard component of the core methods in this research? Note that if the LLM is used only for writing, editing, or formatting purposes and does \emph{not} impact the core methodology, scientific rigor, or originality of the research, declaration is not required.
    \item[] Answer: \answerNA{}
    \item[] Justification: The core method development in this research does not involve LLMs as any important, original, or non-standard components.

\end{enumerate}